\documentclass[5pt]{article}

\usepackage[letterpaper]{geometry}
\usepackage{spconf,amsmath,epsfig}
\usepackage{enumitem}
\usepackage[utf8]{inputenc}
\usepackage[english]{babel}
\usepackage{url}
\usepackage{hyperref}
\usepackage[compact]{titlesec}
\usepackage{fancyhdr}
\begin{document}
\def\x{{\mathbf x}}
\def\L{{\cal L}}
\title{Deep image representations using caption generators}
%
\name{Konda Reddy Mopuri, Vishal B. Athreya  and R. Venkatesh Babu}
\address{ Video Analytics Lab, Computational and Data Sciences\\
	Indian Institute of Science, India\\
\tt\small sercmkreddy@grads.cds.iisc.ac.in, bvishal.athreya@gmail.com, venky@cds.iisc.ac.in}

\maketitle

\begin{abstract}
Deep learning exploits large volumes of labeled data to learn powerful models. When the target dataset is small, it is a common practice to perform transfer learning using pre-trained models to learn new task specific representations. However, pre-trained CNNs for image recognition are provided with limited information about the image during training, which is label alone. Tasks such as scene retrieval suffer from features learned from this weak supervision and require stronger supervision to better understand the contents of the image. In this paper, we exploit the features learned from caption generating models to learn novel task specific image representations. In particular, we consider the state-of-the art captioning system Show and Tell~\cite{SnT-pami-2016} and the dense region description model DenseCap~\cite{densecap-cvpr-2016}. We demonstrate that, owing to richer supervision provided during the process of training, the features learned by the captioning system perform better than those of CNNs. Further, we train a siamese network with a modified pair-wise loss to fuse the features learned by~\cite{SnT-pami-2016} and~\cite{densecap-cvpr-2016} and learn image representations suitable for retrieval. Experiments show that the proposed fusion exploits the complementary nature of the individual features and yields state-of-the art retrieval results on benchmark datasets.
\end{abstract}

\section{Introduction}
\label{sec:intro}
Deep learning has enabled us to learn various sophisticated models using large amounts of labeled data. Computer vision tasks such as image recognition, segmentation, face recognition, etc. require large volumes of labeled data to build reliable models. However, when the training data is not sufficient, in order to avoid over-fitting, it is a common practice to use pre-trained models rather than training from scratch. This enables us to utilize the large volumes of data (eg:~\cite{imagenetmini-ijcv-2015}) on which the pre-trained models are learned and transfer that knowledge to the new target task. Hierarchical nature of the learned representations and task specific optimization makes it easy to reuse them.
This process of reusing  pre-training and learning new task specific representations is referred to as transfer learning or fine-tuning the pre-trained models. There exist many successful instances of transfer learning~(e.g.~\cite{deeplab-iclr-2015}) in computer vision using Convolution Neural Networks (CNNs). Large body of these adaptations are fine-tuned architectures of the well-known recognition models~\cite{deepcnn-nips-2012,vgg-arxiv-2014,googlenet-cvpr-2015,resnet-2015,places-nips-2014,objectdetectors-iclr-2015} trained on the IMAGENET~\cite{imagenetmini-ijcv-2015} and Places~\cite{places-nips-2014} datasets. 

\begin{figure}[t]
\includegraphics[width=0.475\textwidth]{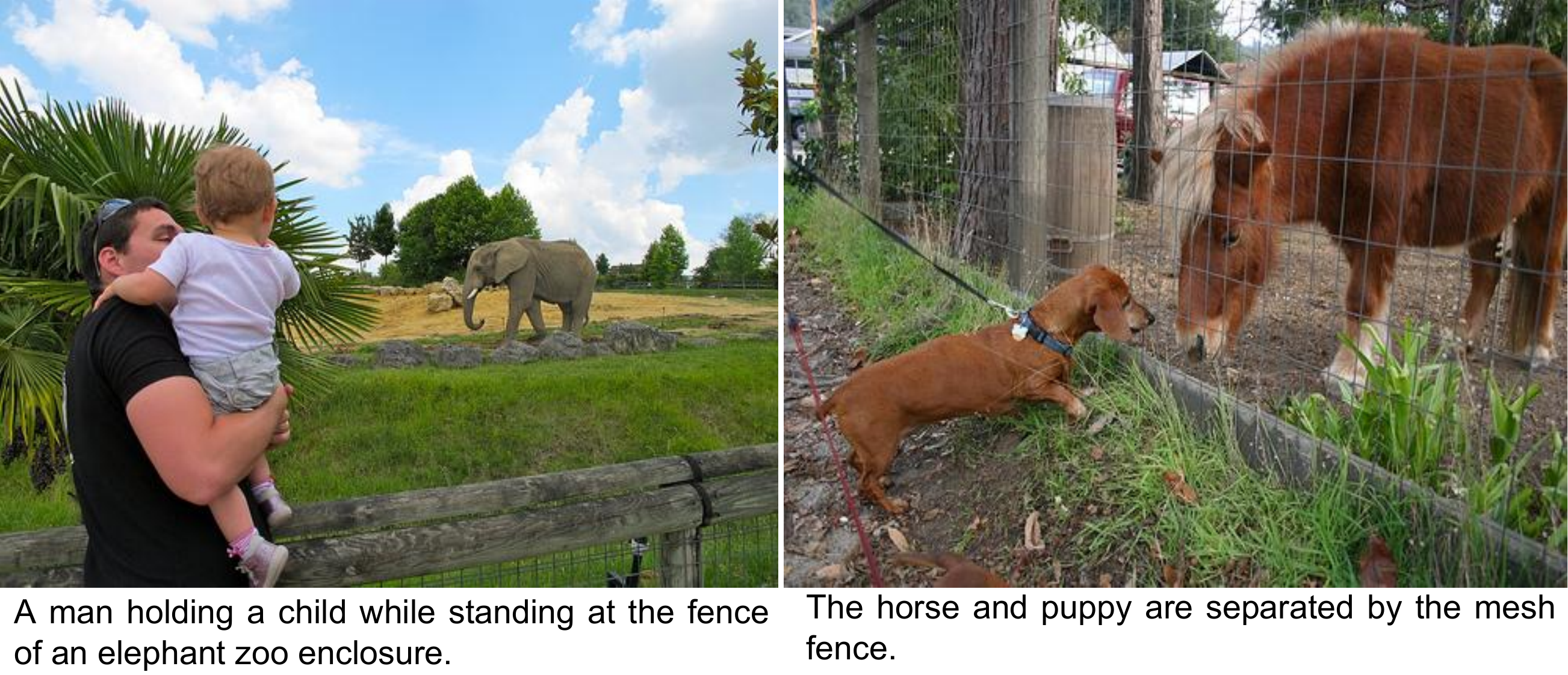}
\caption{Sample image-caption pairs from MSCOCO dataset~\cite{mscoco-eccv-2014}. The caption gives more information than just listing down the objects present in the image. For example, caption for image on the right panel provides additional information such as \textit{separated by mesh fence} which is not available with the labels alone. The proposed method exploits the features learned with this stronger supervision to achieve efficient task specific image representations.}
\label{fig:fig1}
\end{figure}

\begin{figure}[t]
\includegraphics[width=0.5\textwidth]{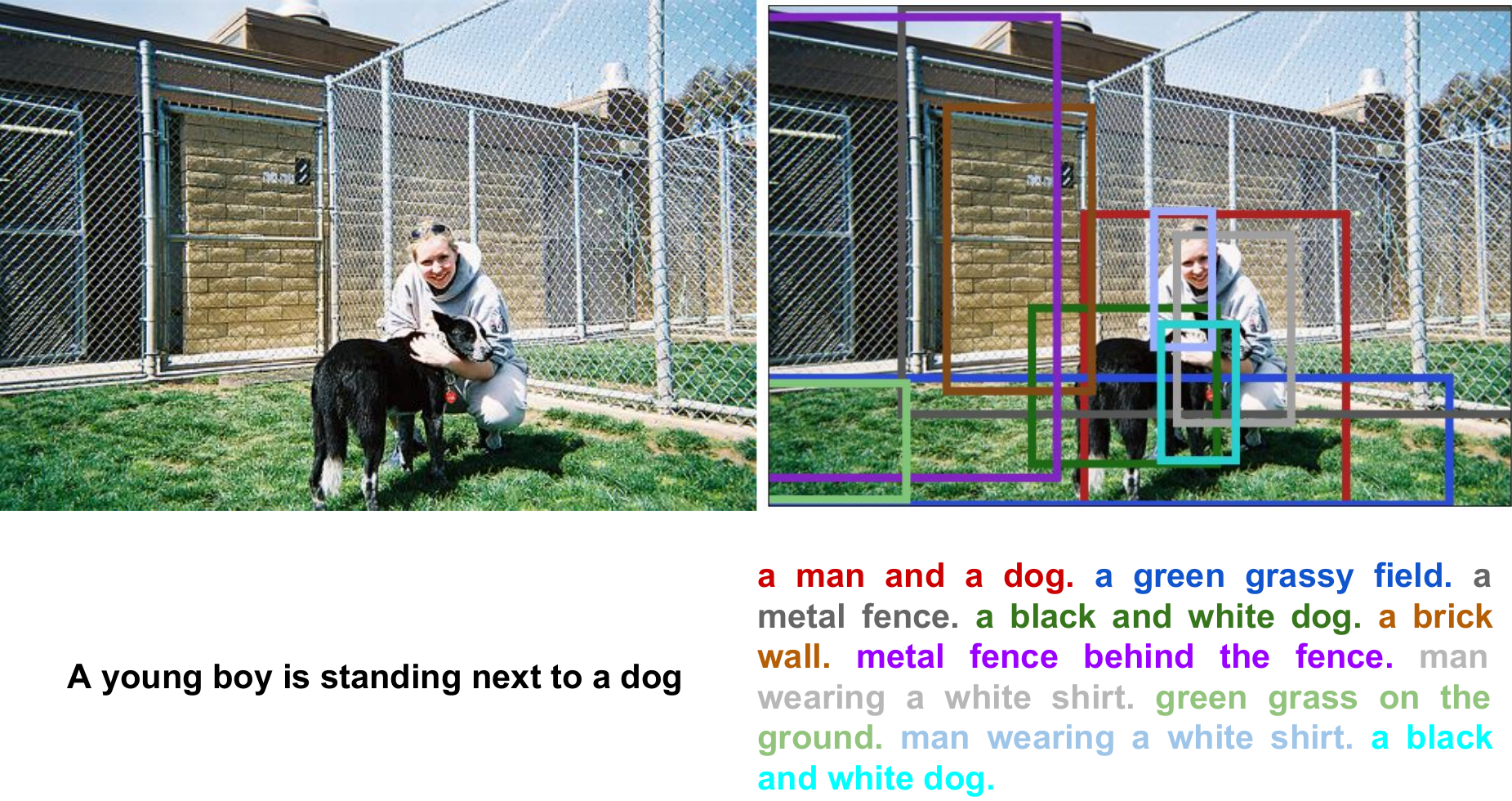}
\caption{Predictions of Show and Tell~\cite{SnT-pami-2016} and Densecap~\cite{densecap-cvpr-2016} models on a sample image from the rPascal~\cite{attributegraph-iccv-2015} dataset. Note that they are complementary in nature: \cite{SnT-pami-2016} (left panel) gives summary of the scene, where as \cite{densecap-cvpr-2016} (right panel) provides more details about the individual objects. The Proposed fusion network benefits from both to learn an efficient image representation.}
\label{fig:comp-1}
\end{figure}

However, these models perform object or scene classification and have very limited information about the image. All that these models are provided with during training is the category label. No other information about the scene is provided. For example, the image shown in Figure \ref{fig:comp-1} has \emph{dog and person} as labels. Useful information such as \emph{indoor or outdoor, interaction between the objects, presence of other objects in the scene} is missing. Tasks such as image search (similar image retrieval) suffer from the features learned by this weak supervision. Image retrieval requires the models to understand the image contents in a better manner (eg:~\cite{oldf,img-text}) to be able to retrieve similar images. Specifically, when the images have multiple objects and graded relevance scores (multiple similarity levels, eg: on a scale from $1$ to $5$) instead of binary relevance (similar or dissimilar), the problem becomes more severe.

On the other hand, automatic caption generation models~(e.g.~\cite{SnT-cvpr-2015,SnT-pami-2016, neuraltalk-cvpr-2015, mrnn-iclr-2015}) are trained with human given descriptions of the images. These models are trained with stronger supervision compared to the recognition models. For example, Figure~\ref{fig:fig1} shows pair of images form MSCOCO~\cite{mscoco-eccv-2014} dataset along with their captions. Richer information is available to these models about the scene than mere labels. In this paper, we exploit the features learned via strong supervision by these models and learn task specific image representations for retrieval via pairwise constraints. 

In case of CNNs, the learning acquired from training for a specific task (e.g. recognition on IMAGENET) is transferred to other vision tasks. Transfer learning followed by task specific fine-tuning has proven to be efficient to tackle less data scenarios. However, similar transfer learning is left unexplored in the case of caption generators. For the best of our knowledge, this is the first attempt to explore that knowledge via fine-tuning the representations learned by them to a retrieval task.

The major contributions of our work can be listed as:
\begin{itemize}
    \item We show that the features learned by the image captioning systems represent image contents better than those of CNNs via image retrieval experiments. We attempt to exploit the strong supervision observed during their training via transfer learning. 
    \item We train a siamese network using a modified pair-wise loss suitable for non-binary relevance scores to fuse the complementary features learned by \cite{SnT-pami-2016} and \cite{densecap-cvpr-2016}. We demonstrate that the task specific image representations learned via our proposed fusion achieve state-of-the-art performance on benchmark retrieval datasets.    
\end{itemize}
The paper is organised as follows: Section~\ref{sec:pa} provides a short summary of \cite{SnT-pami-2016} and \cite{densecap-cvpr-2016} before presenting details about the proposed approach to perform transfer learning. This section also discusses the proposed fusion architecture. Section~\ref{sec:expts} details the experiments performed on benchmark datasets and discusses various aspects along with the results. 
Finally, Section~\ref{sec:conclu} concludes the paper. 

\section{Alternate Image Representations}
\label{sec:pa}
Transfer learning followed by task specific fine-tuning is a well known technique in deep learning. In this section we present an approach to exploit the fine supervision employed by the captioning models and the resulting features. Especially, we target the task of similar image retrieval and learn suitable features. 

\begin{figure}[t]
\includegraphics[width=0.475\textwidth]{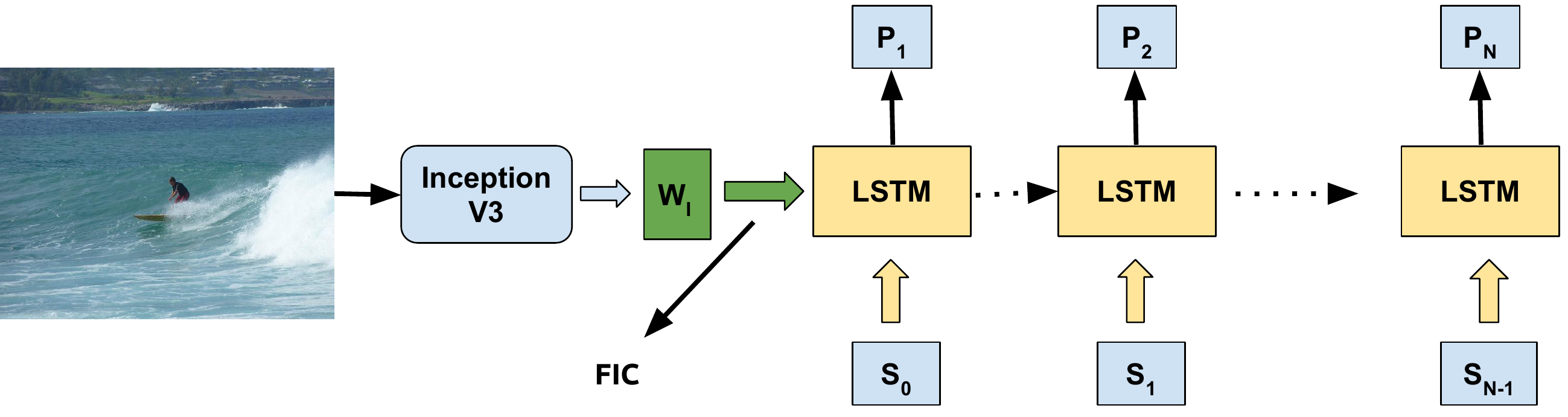}
\caption{Overview of the Show and Tell~\cite{SnT-pami-2016} system. Note that the green arrow indicates the image encodings and the orange denotes word embeddings. $S_i$ is the predicted word at time instant $i$ and $P_i$ denotes the output of the soft-max probability distribution over the dictionary words.}
\label{fig:SnT-overview}
\end{figure}

Throughout the paper, we consider the state-of-the art captioning model \emph{Show and Tell} by Vinyals \textit{et al.}~\cite{SnT-pami-2016}. Their model is an encoder-decoder framework containing a simple cascading of a CNN to an LSTM. The CNN encodes visual information from the input image and feeds via a learnable transformation $W_I$ to the LSTM. This is called image encoding, which is shown in Figure~\ref{fig:SnT-overview} in green color. The LSTM's task is to predict the caption word by word conditioned on the image and previous words. Image encoding is the output of a transformation ($W_I$) learned from the final layer of the CNN (Inception V3 \cite{inceptionv3-cvpr-2016}) before it is fed to the LSTM. The system is trained end-to-end with image-caption pairs to update the image and word embeddings along with the LSTM parameters. Note that the Inception V3 layers (prior to image encoding) are frozen (not updated) during the first phase of training and they are updated during the later phase.

The features at the image encoding layer $W_I$ (green arrow in Figure~\ref{fig:SnT-overview}) are learned from scratch. Note that these are the features input to the text generating part and fed only once. 
These features are very effective to summarize all the important visual content in the image to be described in the caption. These features need to be more expressive than the deep fully connected layers of the typical CNNs trained with weak supervision (labels). Therefore, we consider transferring these features to learn task specific features for image retrieval. We refer to these features as \emph{Full Image Caption} $(FIC)$ features since the generated caption gives a visual summary of the whole image.

On the other hand Johnson \textit{et al.}~\cite{densecap-cvpr-2016} proposed an approach to densely describe the regions in the image, called \emph{dense captioning} task. Their model contains a fully convolutional CNN for object localization followed by an RNN to provide the description. Both the modules are linked via a non-linear projection (layer), similar to \cite{SnT-pami-2016}. The objective is to generalize the task of object detection and image captioning. Their model is trained end-to-end over the Visual genome~\cite{visualgenome-arxiv-2016} dataset which provides object level annotations and corresponding descriptions. They fine-tune the later (from fifth) layers of the CNN module (VGG~\cite{vgg-arxiv-2014} architecture) along with training the image encodings and RNN parameters. Similar to $FIC$ features we consider the image encodings to transfer the ability of this model to describe regions in the image. This model provides encodings for each of the described image regions and associated priorities. Figure~\ref{fig:comp-1} (right panel) shows an example image and the region descriptions predicted by \emph{DenseCap} model. Note that the detected regions and corresponding descriptions are dense and reliable. In order to have a summary of the image contents, we perform mean pooling on the representations (features) belonging to top-K (according to the predicted priorities) regions. We refer to the pooled encodings as \emph{Densecap} features.

Especially for tasks such as image retrieval, models trained with strong object and attribute level supervision can provide better pre-trained features than those of weak label level supervision. Therefore, we propose an approach to exploit the \emph{Densecap} features along with the $FIC$ features and learn task specific image representations.

\subsection{Complementary features and Fusion}
\label{sec:comp}
Figure~\ref{fig:comp-1} shows descriptions predicted by \cite{SnT-pami-2016} and \cite{densecap-cvpr-2016} for a sample image. Note that the predictions are complementary in nature. $FIC$ provides the summary of the scene: \emph{a boy is standing next to a dog}. Where as, \emph{Densecap} provides more details about the scene and objects: presence of \emph{green grass, metal fence, brick wall} and attributes of objects such as \emph{black dog, white shirt,}etc. 

In the proposed approach, we attempt to learn image representations that exploit the strong supervision available from the training process of \cite{SnT-pami-2016} and \cite{densecap-cvpr-2016}. Further, we take advantage of the complementary nature of these two features and fuse them to learn task specific features for image retrieval. 
\begin{figure}[t]
\includegraphics[width=0.47\textwidth]{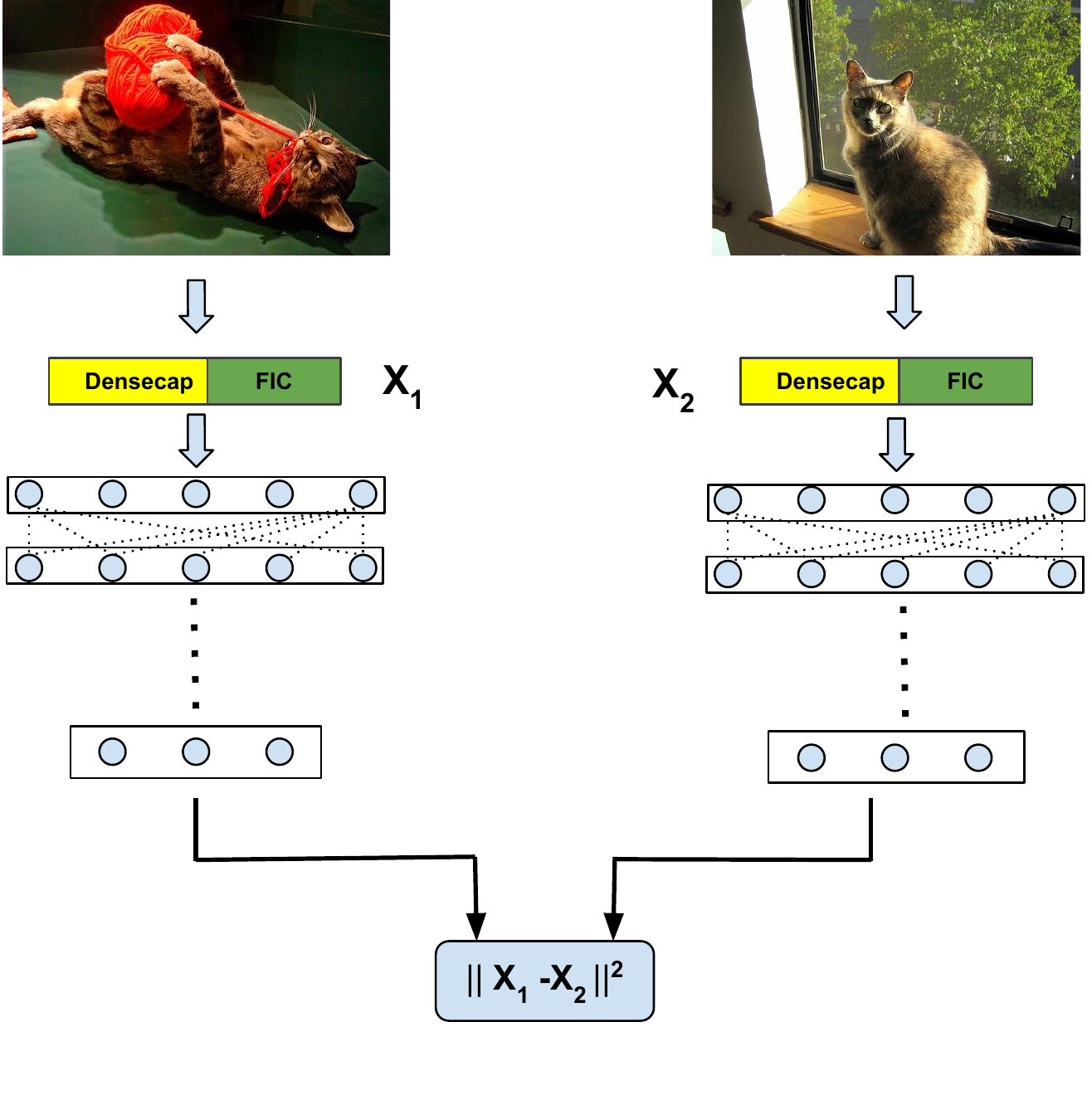}
\caption{The proposed siamese architecture learns image representations by fusing the complementary information extracted from \cite{SnT-pami-2016} and \cite{densecap-cvpr-2016}. Note that both the wings contain a series of dense layers whose weights are tied. The similarity is computed at the final layer and error is back propagated to update the intermediate layers.}
\label{fig:siamese}
\end{figure}
We train a siamese network to fuse both the features. The overview of the architecture is presented in Figure~\ref{fig:siamese}. The proposed siamese architecture has two wings. A pair of images is presented to the network along with their relevance score (high for similar images, low for dissimilar ones). In the first layer of the architecture, $FIC$ and \emph{Densecap} features are late fused (concatenated) and presented to the network. A sequence of layers is added on both the wings to learn discriminative embeddings. Note that these layers on both the wings have tied weights (identical transformations in the both the paths). In the final layer, the features are compared to find the similarity and the loss is computed with respect to the ground truth relevance. The error gets back-propagated to update the network parameters. Our network accepts the complementary information provided by both the features and learns a metric via representations suitable for image retrieval. More details about the training are presented in section~\ref{subsec:training}.

\section{Experiments}
\label{sec:expts}
\subsection{Datasets}
\begin{figure*}[th]
\includegraphics[width=\textwidth]{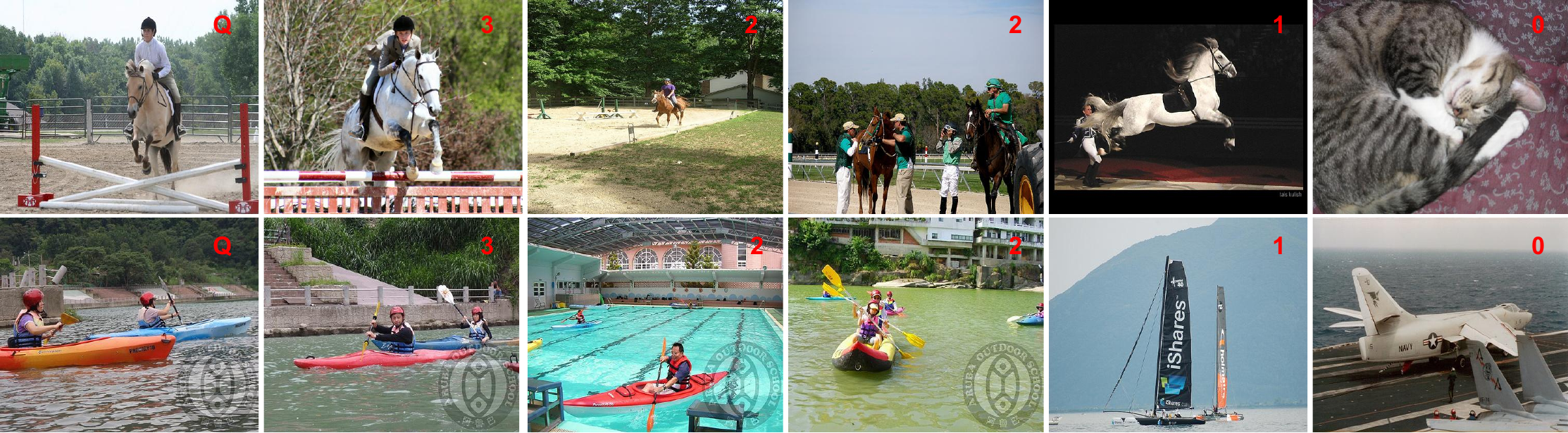}
\caption{Example query images and their relevant images. Top row belongs to rPascal and bottom row belongs to rImagenet dataset. Note that in the top right corner of the image it's relevance to query is shown in red and images are arranged in decreasing order of their relevance.}
\label{fig:sample-db-images}
\end{figure*}
\begin{figure}[t]
\includegraphics[width=0.5\textwidth]{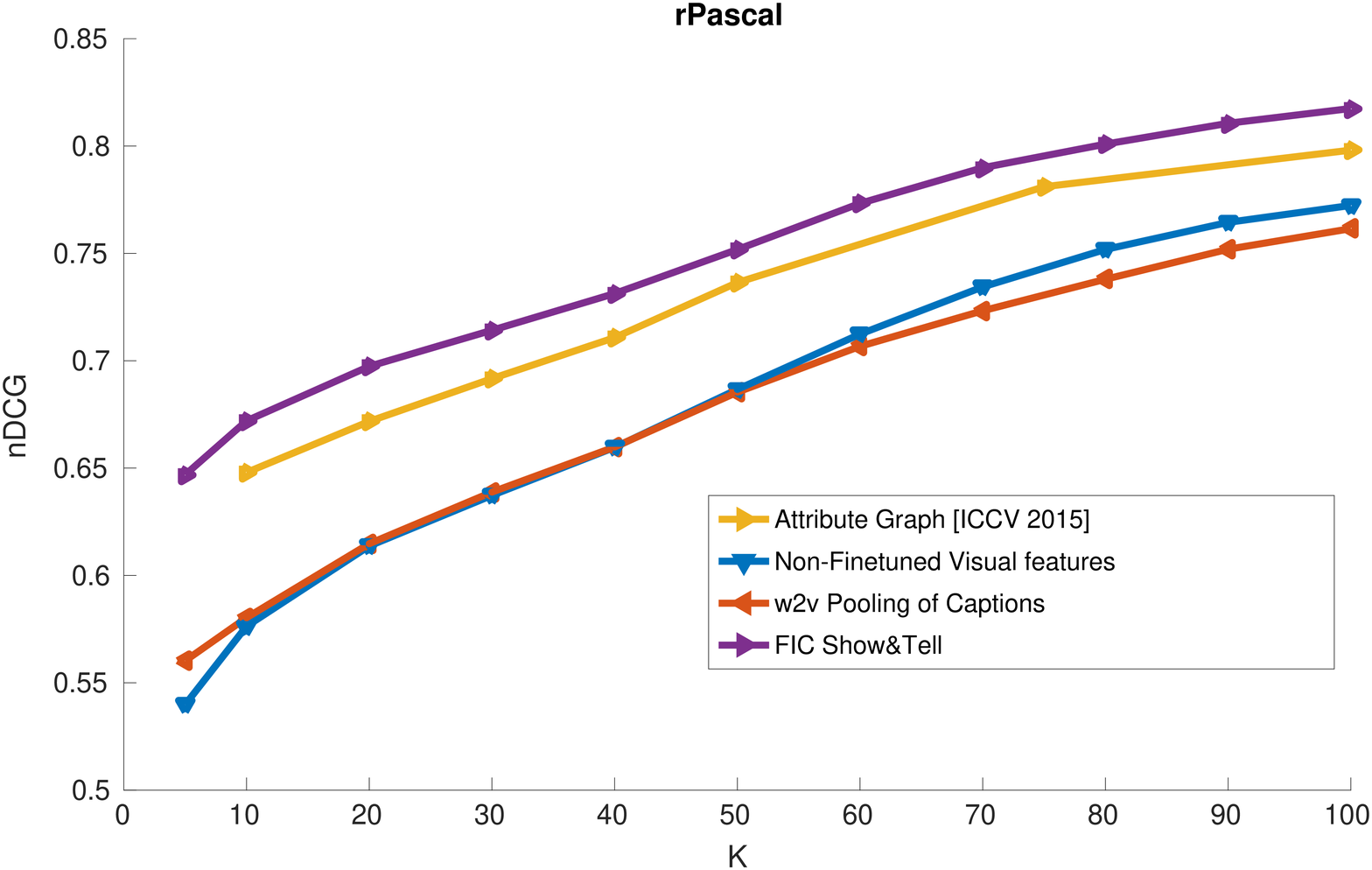}
\caption{Retrieval performance of the $FIC$ features on rPascal dataset. Note that the $FIC$ features learned by the caption generation system~\cite{SnT-pami-2016} outperform the underlying inception v3 features (non-finetuned visual features) and the word2vec pooled representations of the predicted captions. They also beat the state-of-the art Attribute Graph method~\cite{attributegraph-iccv-2015}.}
\label{fig:pascal-comp}
\end{figure}
\vspace{-0.15cm}
\begin{figure}[t]
\includegraphics[width=0.5\textwidth]{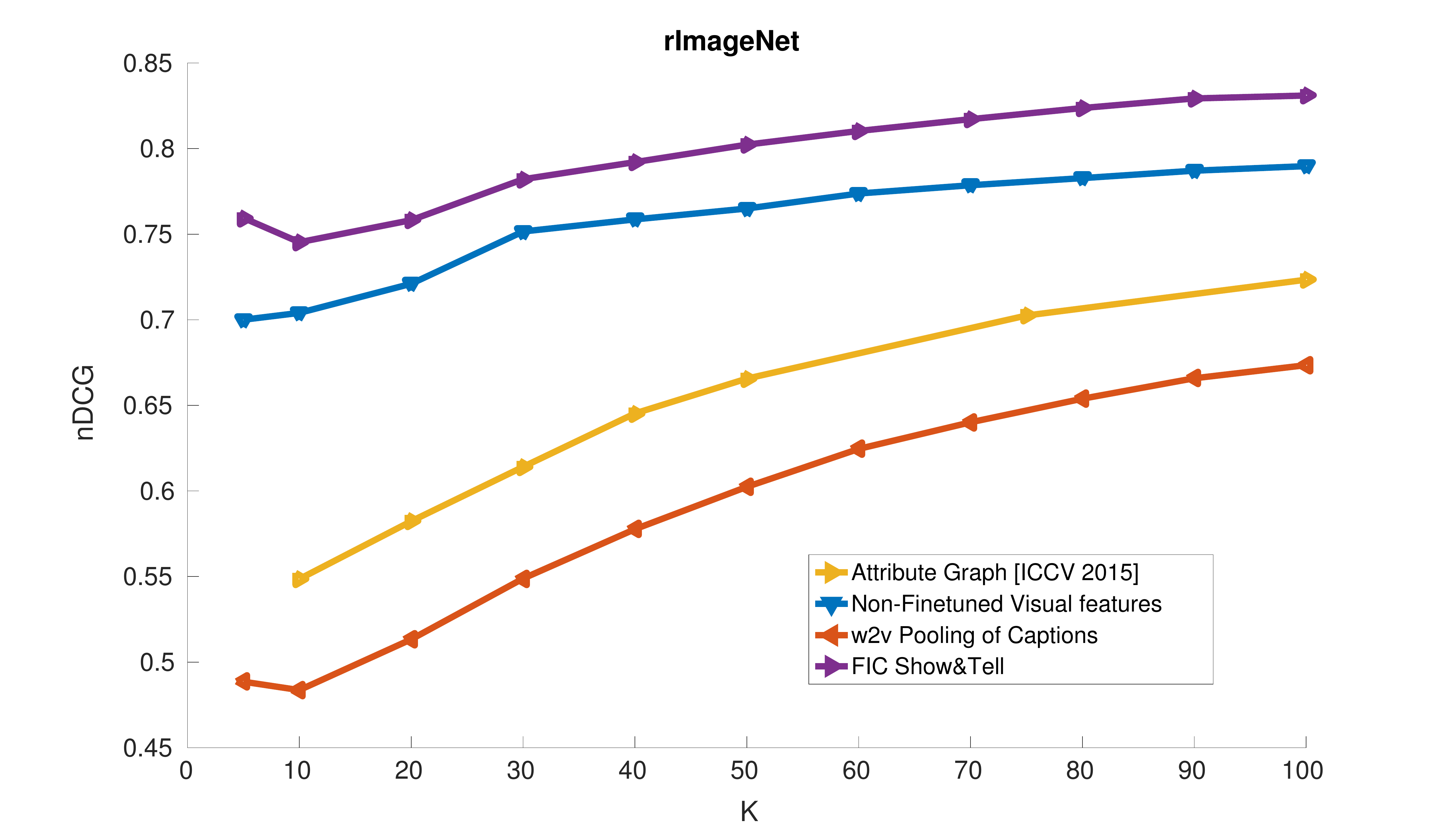}
\caption{Retrieval performance of the $FIC$ features on rImagenet dataset. Note that the $FIC$ features learned by the caption generation system~\cite{SnT-pami-2016} outperform the underlying inception v3 features (non-finetuned visual features) and the word2vec pooled representations of the predicted captions. They also beat the state-of-the art Attribute Graph method~\cite{attributegraph-iccv-2015}.}
\label{fig:imagenet-comp}
\end{figure}
We begin by explaining the retrieval datasets\footnote{The datasets are available at \url{http://val.serc.iisc.ernet.in/attribute-graph/Databases.zip}} considered for our experiments. In order to have more natural scenario, we consider retrieval datasets that have graded relevance scores instead of binary relevance (similar or dissimilar). We require the relevance to be assigned based on overall visual similarity as opposed to any one particular aspect of the image (e.g. objects). To match these requirements, we consider two datsets \emph{rPascal} (ranking Pascal) and \emph{rImagenet} (ranking Imagenet) composed by Prabhu \textit{et al.}~\cite{attributegraph-iccv-2015}. These datasets are subsets of aPascal~\cite{apascal-cvpr-2009} and Imagenet~\cite{imagenetmini-ijcv-2015} respectively. Each of the datasets contains $50$ query images and a set of corresponding relevant images. They are composed by $12$ annotators participating to assign relevance scores. The scores have $4$ grades, ranging from $0$ (irrelevant) to $3$ (excellent match).

\begin{itemize}
\item{rPascal}:
This dataset is composed from the test set of aPascal~\cite{apascal-cvpr-2009}. The queries comprise of $18$ indoor and $32$ outdoor scenes. The dataset consists of a total of $1835$ images with an average of $180$ reference images per query.
\item{rImagenet}:
It is composed from the validation set of ILSVRC 2013 detection challenge. Images containing at least $4$ objects are chosen. The queries contain $14$ indoor scenes and $36$ outdoor scenes. The dataset consists of a total of $3354$ images with an average of $305$ reference images per query.
\end{itemize}

Figure~\ref{fig:sample-db-images} shows sample images from the two datasets. Note that the first image in each row is query and the following images are reference images with relevance scores displayed at top right corner.

\subsection{Evaluation metric}
We followed the evaluation procedure presented in \cite{attributegraph-iccv-2015}. For quantitative evaluation of the performance, we compute normalized Discounted Cumulative Gain (nDCG) of the retrieved list. nDCG is a standard evaluation metric used for ranking algorithms~(e.g. \cite{ndcg-ex} and \cite{attributegraph-iccv-2015}). For all the queries in each dataset, we find the nDCG value and report the mean nDCG per dataset evaluated at different ranks (K).

\subsection{Features learned by the caption generator ($FIC$)}
\begin{figure}[t]
\includegraphics[width=0.5\textwidth]{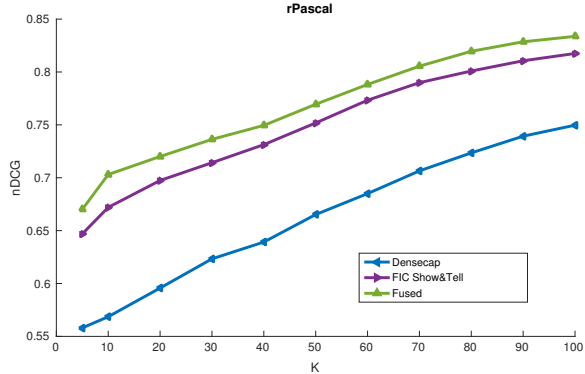}
\caption{Retrieval performance of the learned image representations over rPascal dataset.}
\label{fig:pascal-fused}
\end{figure}
In this subsection we demonstrate the effectiveness of the features obtained from the caption generation model~\cite{SnT-pami-2016}. For each image we extract the $512D$ $FIC$ features to encode it's contents. Note that these are the features learned by the caption generation model via the strong supervision provided during the training. Retrieval is performed by computing distance between the query and the reference images' features and arranging in the increasing order of the distances. 

Figure~\ref{fig:pascal-comp} and ~\ref{fig:imagenet-comp} show the plots of nDCG evaluted at different ranks $(K)$ on the two datasets. As a baseline comparison, we have compared the performance of $FIC$ features with that of the non-finetuned visual features of inception v3 model (blue color plot in Figure~\ref{fig:pascal-comp} and~\ref{fig:imagenet-comp}). Note that these are $2048D$ features that are extracted from the last fully connected layer of the inception v3 model~\cite{inceptionv3-cvpr-2016}. The $FIC$ features outperform the non-finetuned visual features by a large margin emphasizing the effectiveness of the strong supervision.

We have considered another baseline using the natural language descriptors. For a given image, we have predicted the text caption using the Show and Tell~\cite{SnT-pami-2016} model. After pre-processing (stop word removal and lemmatizing), we encode each of the remaining words using word2vec~\cite{word2vec-arxiv-2013} embeddings and mean pool them to form an image representation. Note that the $FIC$ features perform better than this baseline also.

We also compare the performance of $FIC$ features against the state-of-the art Attribute graph approach~\cite{attributegraph-iccv-2015}. The $FIC$ features clearly outperform the Attribute Graph approach in case of both the benchmark datasets.

\subsection{Task specific fine-tuning through fusion}
\label{subsec:training}
\begin{figure}[t]
\includegraphics[width=0.5\textwidth]{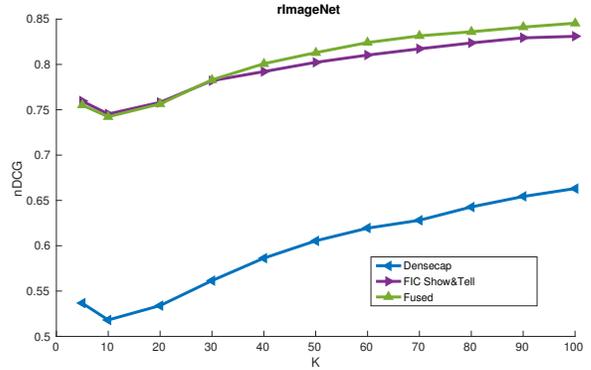}
\caption{Retrieval performance of the learned image representations over rImagenet dataset.}
\label{fig:imagenet-fused}
\end{figure}
The proposed fusion architecture\footnote{Project codes can be found at \url{https://github.com/mopurikreddy/strong-supervision}} is trained with pairs of images and corresponding relevance scores $(y)$. The typical pairwise training consists of binary relevance scores: simila r$(1)$ or dissimilar $(0)$. The objective is to reduce the distance between the projections of the images if they are similar and separate them if dissimilar. Equation~(\ref{eq:typical-siamese-loss}) shows the contrastive loss~\cite{contrastiveloss-ex} typically used to train siamese networks. 

\begin{equation}
E = \frac{1}{2N} \sum\limits_{n=1}^N \left(y\right) d + \left(1-y\right) \max \left(\nabla-d, 0\right)
\label{eq:typical-siamese-loss}
\end{equation}
where, $E$ is the prediction error, $N$ is the mini-batch size, $y$ is the relevance score ($0$ or $1$), $d$ is the distance between the projections of the pair of images and $\nabla$ is the margin to separate the projections corresponding to dissimilar pair of images.

However, in practice images can have non-binary relevance scores. To handle more fine grained relevances, we modified the contrastive loss function to include non-binary scores as shown in equation~(\ref{eq:modified-siamese-loss})

\begin{equation}
E = \frac{1}{2N} \sum\limits_{n=1}^N (y^2) d +  \textbf{1}(y=0) \max (\nabla-d^2, 0)
\label{eq:modified-siamese-loss}
\end{equation}

where $\textbf{1}(.)$ is indicator function.
Note that the modified loss function favours the nDCG measure by strongly punishing (due to the square term) the distances between images with higher relevance scores.

We train a siamese network with $5$ fully connected layers on both the wings, with tied weights. The number of units in each wing are $1024-2048-1024-512-512$. The representations learned at the last layer are normalized and euclidean distance is minimized according to Equation~(\ref{eq:modified-siamese-loss}). We divide the queries into $5$ splits to perform $5$ fold validation and report the mean nDCG. That is, each fold contains image pairs of $40$ queries and corresponding reference images for training. The remaining $10$ queries form the evaluation set. On an average, each fold contains $11300$ training pairs for rPascal and $14600$ pairs for rImagenet.

Figures~\ref{fig:pascal-fused} and \ref{fig:imagenet-fused} show the performance of the task specific image representations learned via the proposed fusion. For evaluating the performance of the Densecap~\cite{densecap-cvpr-2016} method, we have mean pooled the encodings corresponding to top-$5$ image regions resulting a $512D$ feature. $FIC$ feature is also $512D$ vector, therefore forming an input of $1024D$ to the network. Note that the transfer learning and fine-tuning through fusion improves the retrieval performance on both the datasets.

\section{Conclusion}
\label{sec:conclu}
In this paper, we have presented an approach to exploit the strong supervision observed in the training of caption generation systems. We have demonstrated that image understanding tasks such as retrieval can benefit from this strong supervision compared to weak label level supervision. Transfer learning followed by task specific fine-tuning is commonly observed in CNN based vision systems. However similar practice is relatively unexplored in the case of these captioning systems. Our approach can potentially open new directions for exploring other sources for stronger supervision and better learning. It can also motivate to tap the juncture of vision and language in order to build more intelligent systems.
\section{Acknowledgement}
This work was supported by Defence Research and Development Organization (DRDO), Government of India.

\bibliographystyle{IEEEbib}
\bibliography{mybibliography}

\end{document}